\documentclass{article} 
\usepackage{tencent_ailab_tech_report}
\usepackage[colorlinks = true,
            linkcolor = blue,
            urlcolor  = blue,
            citecolor = blue,
            anchorcolor = blue]{hyperref}   
\usepackage{microtype}
\usepackage{hyperref}
\usepackage{url}
\usepackage{booktabs}
\usepackage{enumitem}
\usepackage{multicol}
\usepackage{multirow}
\usepackage{CJKutf8}
\usepackage{amsmath}
\usepackage{siunitx}
\usepackage{floatflt}
\usepackage{wrapfig}
\usepackage{graphicx}
\usepackage{booktabs}
\usepackage{wrapfig}
\usepackage{authblk}
\usepackage{lipsum}

\usepackage{algorithm}
\usepackage{algorithmicx}
\usepackage{algpseudocode}
\usepackage{microtype}
\usepackage{graphicx}
\usepackage{subfigure}
\usepackage{multirow}
\usepackage{booktabs} 
\usepackage{pifont}  
\usepackage{graphicx}  
\usepackage{pgfplots}

\usepackage{hyperref}

\usepackage{amssymb} 

\algnewcommand{\LeftComment}[1]{\Statex \(\triangleright\) #1}

\usepackage{array}
\usepackage{amsmath}
\usepackage{amssymb}
\usepackage{mathtools}
\usepackage{amsthm}
\usepackage{arydshln}
\usepackage[capitalize,noabbrev]{cleveref}
\usepackage{adjustbox} 
\usepackage{enumitem}
\usepackage{lineno}
\theoremstyle{plain}

\theoremstyle{definition}

\theoremstyle{remark}

\usepackage[textsize=tiny]{todonotes}

\definecolor{nred}{RGB}{196, 38, 11}
\definecolor{ngreen}{RGB}{18, 141, 21}
\definecolor{nblue}{RGB}{41, 52, 190}
\definecolor{hzw}{RGB}{223, 97, 76}
\definecolor{lt}{RGB}{54, 89, 170}

\newcommand{\ignore}[1]{}

\colmfinalcopy

\title{Don't Get Lost in the Trees: Streamlining LLM Reasoning by Overcoming Tree Search Exploration Pitfalls}

\author[ ]{Ante Wang\thanks{The work was done when Ante Wang was interning at Tencent AI Lab.}~~$^{,1}$}
\author[ ]{\mbox{Linfeng Song}\thanks{Correspondence to: Linfeng Song \textless lfsong@tencent.com\textgreater, Zhaopeng Tu \textless zptu@tencent.com\textgreater, and Jinsong Su \textless jssu@xmu.edu.cn\textgreater.}~~$^{2}$}
\author[ ]{Ye Tian$^{2}$}
\author[ ]{Dian Yu$^{2}$}
\author[ ]{\mbox{Haitao Mi}$^{2}$}
\author[ ]{Xiangyu Duan$^{3}$}
\author[ ]{\\ \mbox{Zhaopeng Tu}$^{\dagger\,2}$}
\author[ ]{\mbox{Jinsong Su}$^{\dagger\,1}$}
\author[ ]{Dong Yu$^{2}$}

\affil[1]{Xiamen University}
\affil[2]{Tencent AI Lab}
\affil[3]{Soochow University}

\newcommand{\method}[0]{\textsc{Fetch}}

\begin{document}

\maketitle

\begin{figure}[h!]
\centering
\vspace{-10mm}
\subfigure[Generated Tokens]{
    \includegraphics[width=0.4\textwidth]{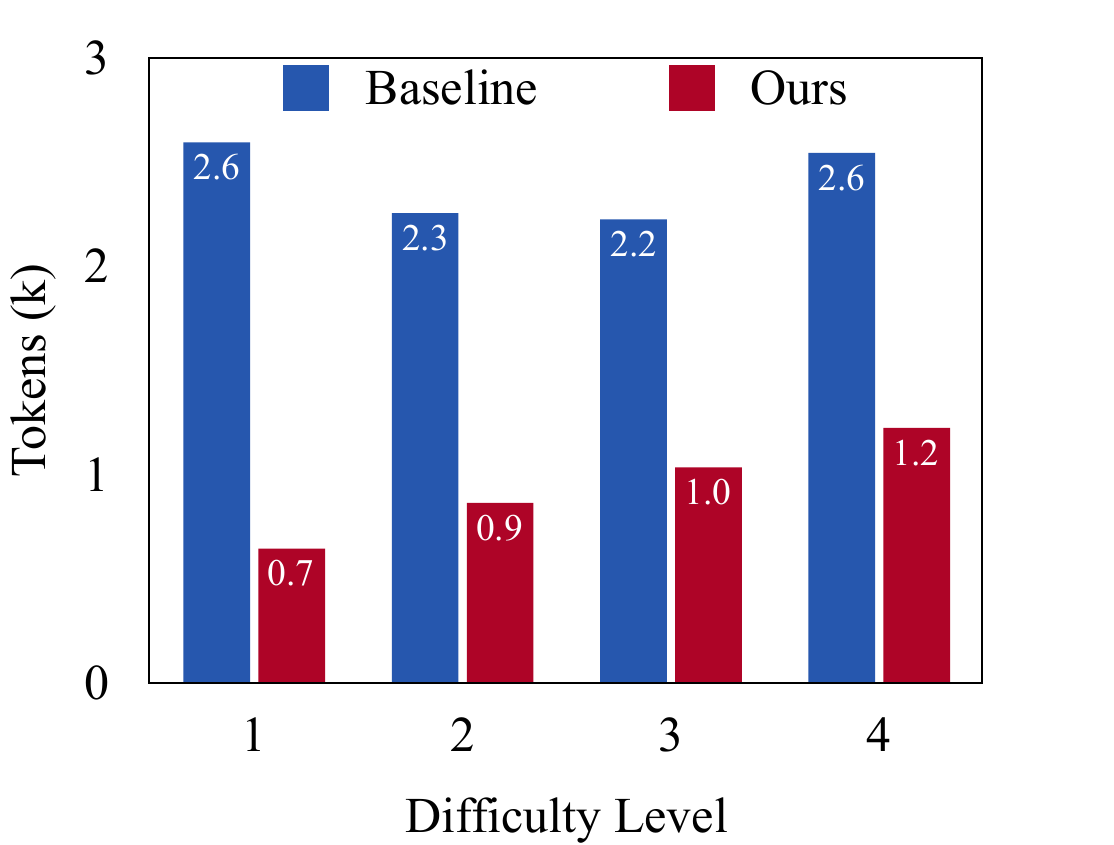}} \hspace{0.05\textwidth}
\subfigure[Accuracy]{
    \includegraphics[width=0.4\textwidth]{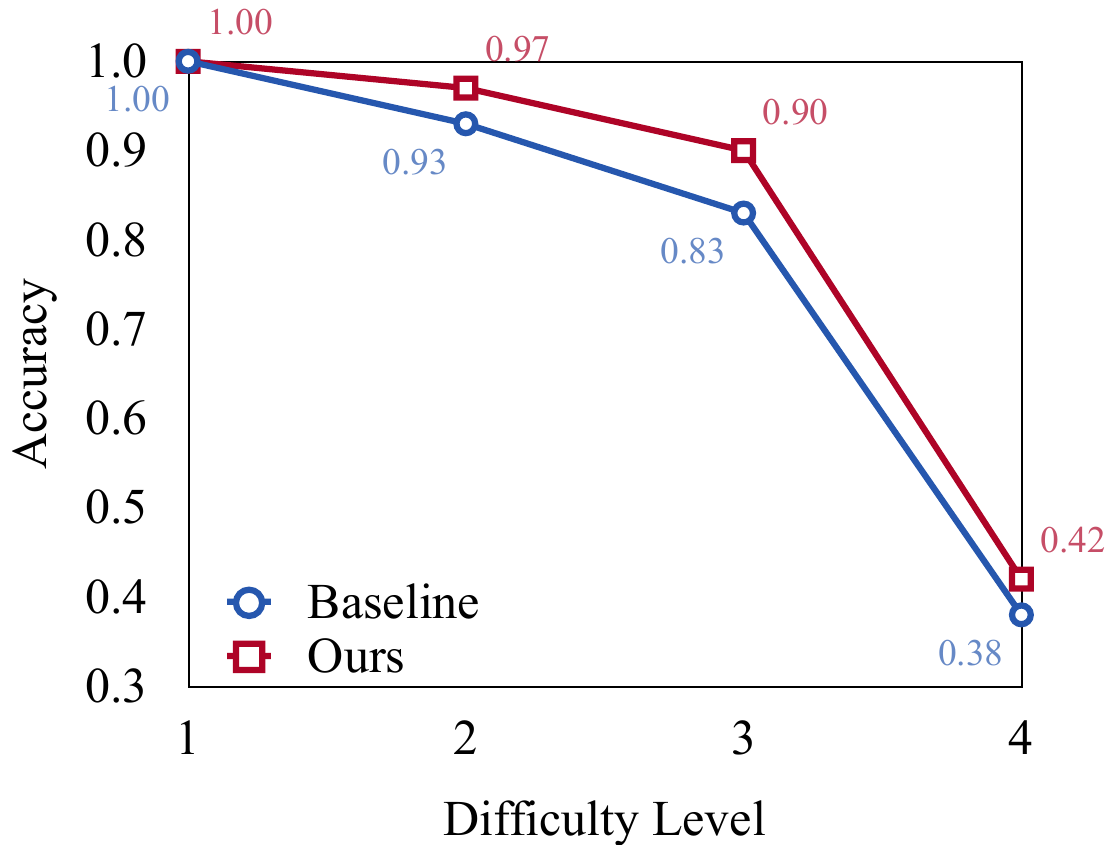}}
\caption{The effect of \emph{\method{}} using Best-First Search as baseline on GSM8K problems of varying difficulty is illustrated. Figure (a) shows that \emph{\method{}} reduces computational costs and prioritizes resource allocation to harder tasks (Levels 3–4) over simpler ones (Levels 1–2), addressing both over-exploration in basic problems and under-exploration in complex cases (a). This leads to better performance, as shown in Figure (b).}
\label{fig:intro}
\end{figure}

\begin{abstract}
Recent advancements in tree search algorithms guided by verifiers have significantly enhanced the reasoning capabilities of large language models (LLMs), but at the cost of increased computational resources. In this work, we identify two key challenges contributing to this inefficiency: \textit{over-exploration} due to redundant states with semantically equivalent content, and \textit{under-exploration} caused by high variance in verifier scoring leading to frequent trajectory switching. 
To address these issues, we propose \textsc{Fetch} -- an  e{\bf f}fici{\bf e}nt {\bf t}ree sear{\bf ch} framework, which is a flexible, plug-and-play system compatible with various tree search algorithms.
Our framework mitigates over-exploration by merging semantically similar states using agglomerative clustering of text embeddings obtained from a fine-tuned SimCSE model. To tackle under-exploration, we enhance verifiers by incorporating temporal difference learning with adjusted $\lambda$-returns during training to reduce variance, and employing a verifier ensemble to aggregate scores during inference. Experiments on GSM8K, GSM-Plus, and MATH datasets demonstrate that our methods significantly improve reasoning accuracy and computational efficiency across four different tree search algorithms, paving the way for more practical applications of LLM-based reasoning. The code is available at \url{https://github.com/Soistesimmer/Fetch}.
\end{abstract}

\section{Introduction}
In recent months, the remarkable reasoning performance of OpenAI-o1 \citep{openai-o1} has sparked significant research interest in enhancing the deductive capabilities of large language models (LLMs, \citealt{touvron2023llama,jiang2023mistral,achiam2023gpt}) through inference-time scaling.
One promising area in this field is the use of tree search algorithms guided by verifiers.
Their effectiveness in finding optimal solutions for complex problems by exploring large search spaces has been demonstrated in various studies~\citep{feng2023alphazero,yao2024tree,yu2024ovm,tian2024toward,kang2024mindstar,wang2024litesearch,zhang2024accessing}.

However, the considerable increase in computational costs presents a significant challenge to the practical applications of these algorithms. Through our pilot study, we have identified the following two main issues that cause this inefficiency:
\begin{itemize}[leftmargin=10pt]
    \item \textit{Over-exploration}: Due to the uncontrollable sampling of reasoning steps from LLMs, the search tree inevitably contains redundant states with semantically equivalent content. 
    This leads to over-exploration of certain reasoning paths, as these redundant states are treated independently.

    \item \textit{Under-exploration}: 
    The verifiers used for search guidance may lack robustness and introduce unnecessary scoring variances due to differences in reasoning style and phrasing choices, particularly when tackling challenging math problems.
    As a result, the search algorithm may frequently switch between different trajectories, leaving many potential paths unfinished by the end of the search.
\end{itemize}

In this paper, we propose \emph{\method{}},  a flexible plug-and-play framework compatible with most existing tree search algorithms to tackle both over and under-exploration issues.
With regard to over-exploration, the framework employs agglomerative clustering to merge similar states in a bottom-up style.
Capturing accurate semantic-level similarity has always been a challenge in natural language processing \citep{dolan2005automatically,agirre2014semeval}, especially when dealing with reasoning trajectories that usually involve complex deductions.
We propose a lightweight sentence embedding model obtained by fine-tuning SimCSE \citep{gao2021simcse} with signals from computationally intensive methods, such as LLM prompting or consistency checking over sampled subsequent steps.
This shares the same spirit as training a reward model to represent expansive human feedback in reinforcement learning from human feedback (RLHF, \citealt{ouyang2022training}).

To tackle under-exploration, we enhance the verifier during both training and inference time to mitigate its variance.
Here we focus on scoring-based verifiers, such as a value network or a process reward model (PRM), as they are widely adopted for tree search \citep{wang2023math,yu2024ovm,tian2024toward,wang2024q}.
One major cause of high variance for such verifiers may be the noisy training signals acquired from Monte Carlo (MC) sampling, as the sample size is small compared to the vast space of possible trajectories.
Drawing inspiration from Temporal Difference (TD) learning \citep{sutton1988learning}, we enhance the verifier during {\em training} by incorporating an adjusted $\lambda$-return. This effectively balances short-term and long-term rewards, thereby reducing the variance.
Furthermore, we propose verifier ensembling during {\em inference}, which aggregates multiple scores from different verifiers. This aggregation effectively averages out the individual biases and variances of each verifier, resulting in a more consistent and reliable estimate.

We conduct experiments on four representative search algorithms: Best-First Search (BFS, \S \ref{sec:bfs}), Beam Search \citep{xie2024self,zhu2024deductive}, Tree Search \citep{wang2024q,wang2024litesearch}, and Monte Carlo Tree Search (MCTS) \citep{feng2023alphazero,tian2024toward}. 
Our experimental results on GSM8K \citep{cobbe2021training}, GSM-Plus \citep{li2024gsm}, and MATH \citep{hendrycks2021measuring} demonstrate that our methods can substantially improve both accuracy and efficiency.
Further analyses reveal that state merging facilitates more efficient search by preventing the over-exploration of redundant states. 
Concerning score variance reduction, both TD($\lambda$) in training and verifier ensembling in inference contribute to generating more accurate and stable scores, with their combination yielding further improvements.

\section{Pilot Study}

This section aims to analyze the over-exploration and under-exploration issues in tree search reasoning.
We first introduce 
definitions and experiment setup,
taking Best-First Search (BFS) as an example (\S \ref{sec:bfs}).
Then, we conduct comprehensive analyses about these two issues on the popular GSM8K \citep{cobbe2021training} benchmark (\S \ref{sec:pilot}).


\subsection{Definitions and Setup}
\label{sec:bfs}
We formulate the search process of a reasoning problem $q$ as a tree search task.
Each tree node (state) $s_i=q, o_1, \dots, o_i$ represents a partial or full reasoning process, where $o_i$ denotes the $i$-th reasoning step.
Most tree search algorithms follow the iteration of two operations: \textit{selection} and \textit{expansion}.
For clarity, we introduce the BFS as a typical example of tree search algorithms.

In the selection phase, BFS selects the most promising node $s_i$ from the set of unexplored nodes $\mathbf{s}$ in the search space based on the feedback from a verifier $v(*)$:
\begin{equation}
    i = \arg \max_{1 \leq j \leq |\mathbf{s}|} v(s_j).
\end{equation}

Then, in the expansion phase, $N$ (expansion size) child nodes are expanded from the selected node by sampling corresponding new reasoning steps from the LLM (or referred to as policy) $\pi(*)$.
Each step is sampled via nucleus sampling \citep{holtzmancurious} with a temperature $\tau$:
\begin{equation}
    o_{i+1} \sim \pi(s_i).
\end{equation}

By iteratively choosing and expanding the most promising node, BFS efficiently navigates toward the goal state, making it particularly effective for optimization problems.

\begin{figure}[t]
    \subfigure[Accuracy]{
    \includegraphics[width=0.24\textwidth]{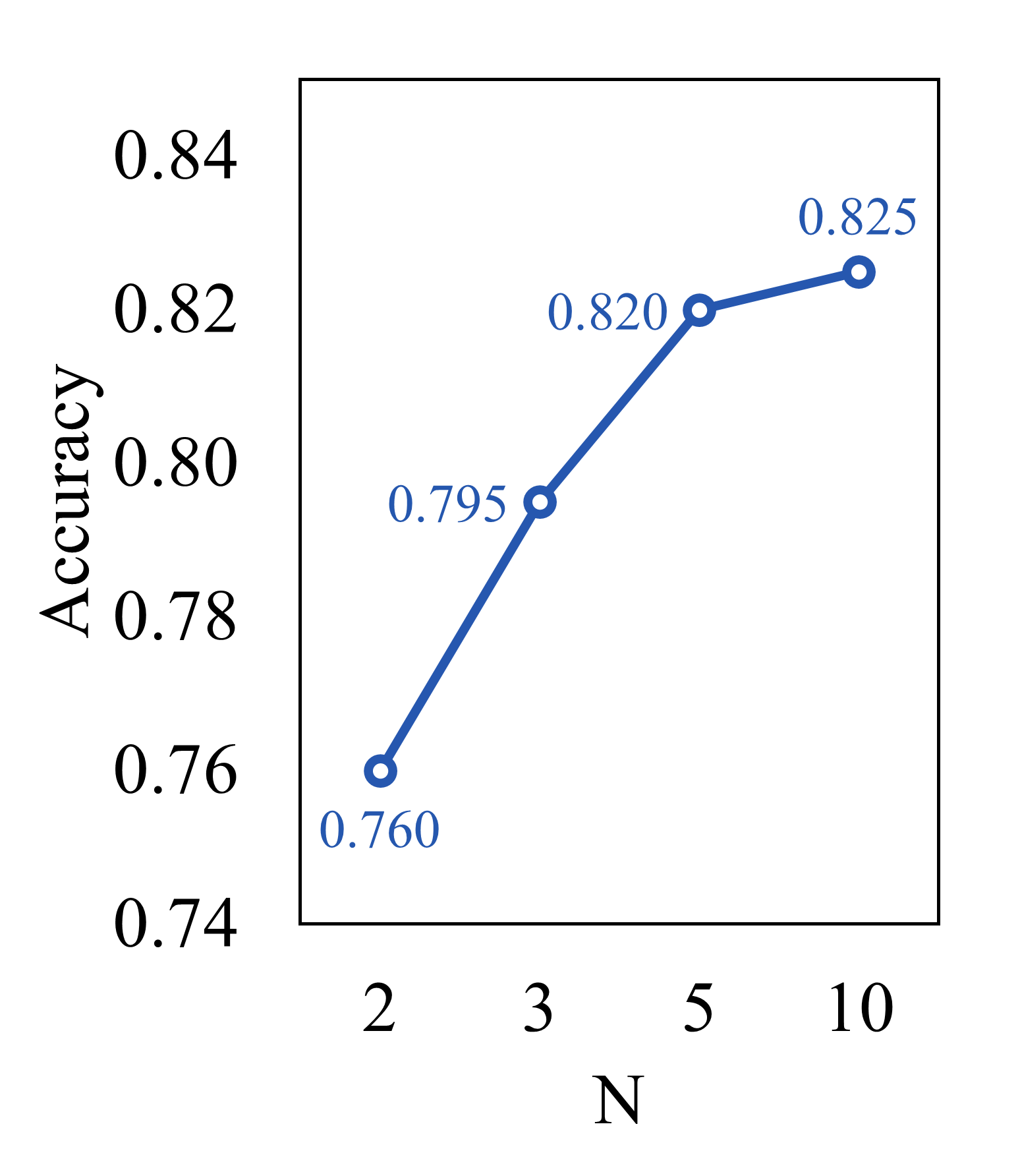}
    }
    \subfigure[Generated Tokens]{
    \includegraphics[width=0.23\textwidth]{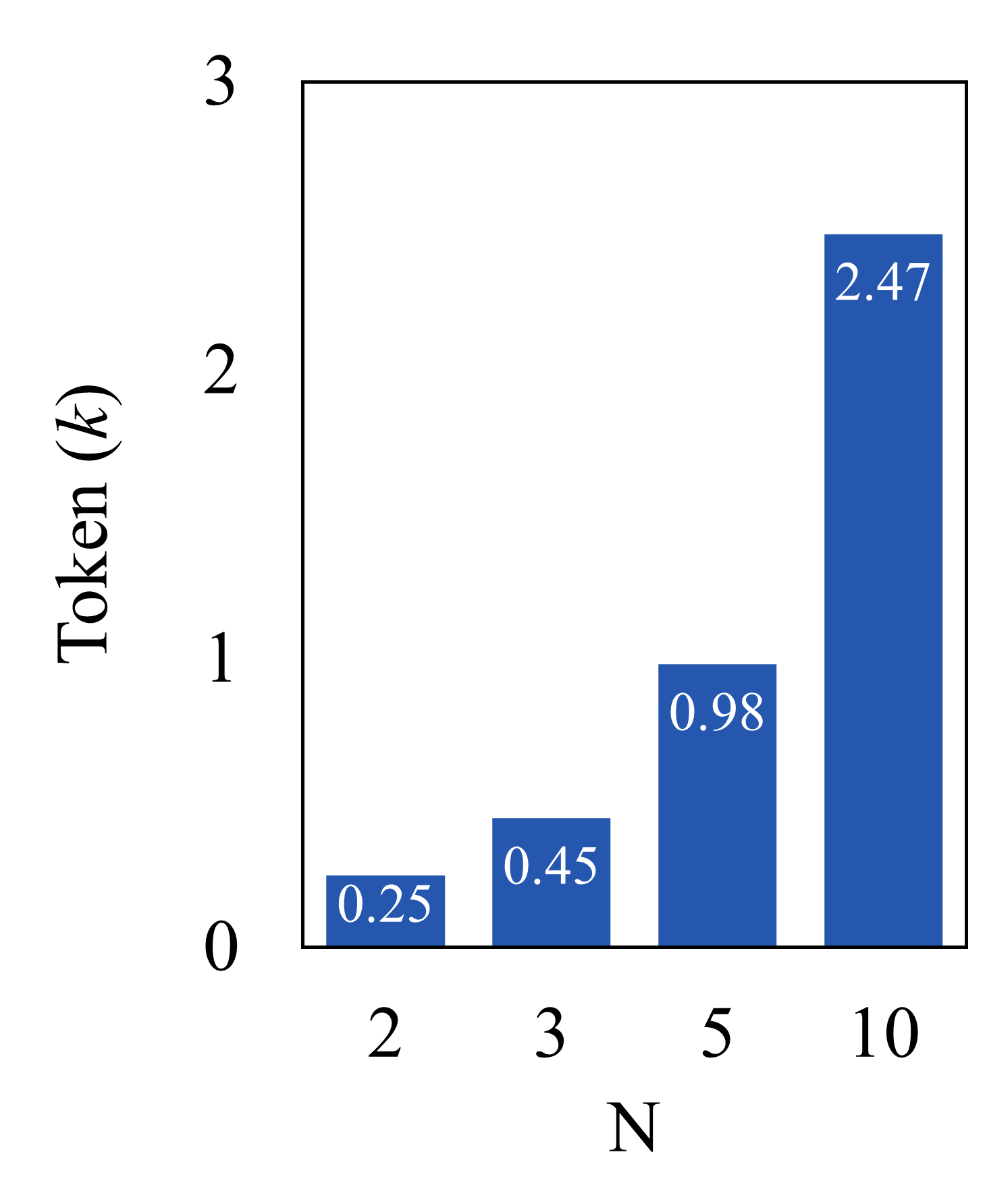}
    }
    \subfigure[Similarity Degree]{
    \includegraphics[width=0.224\textwidth]{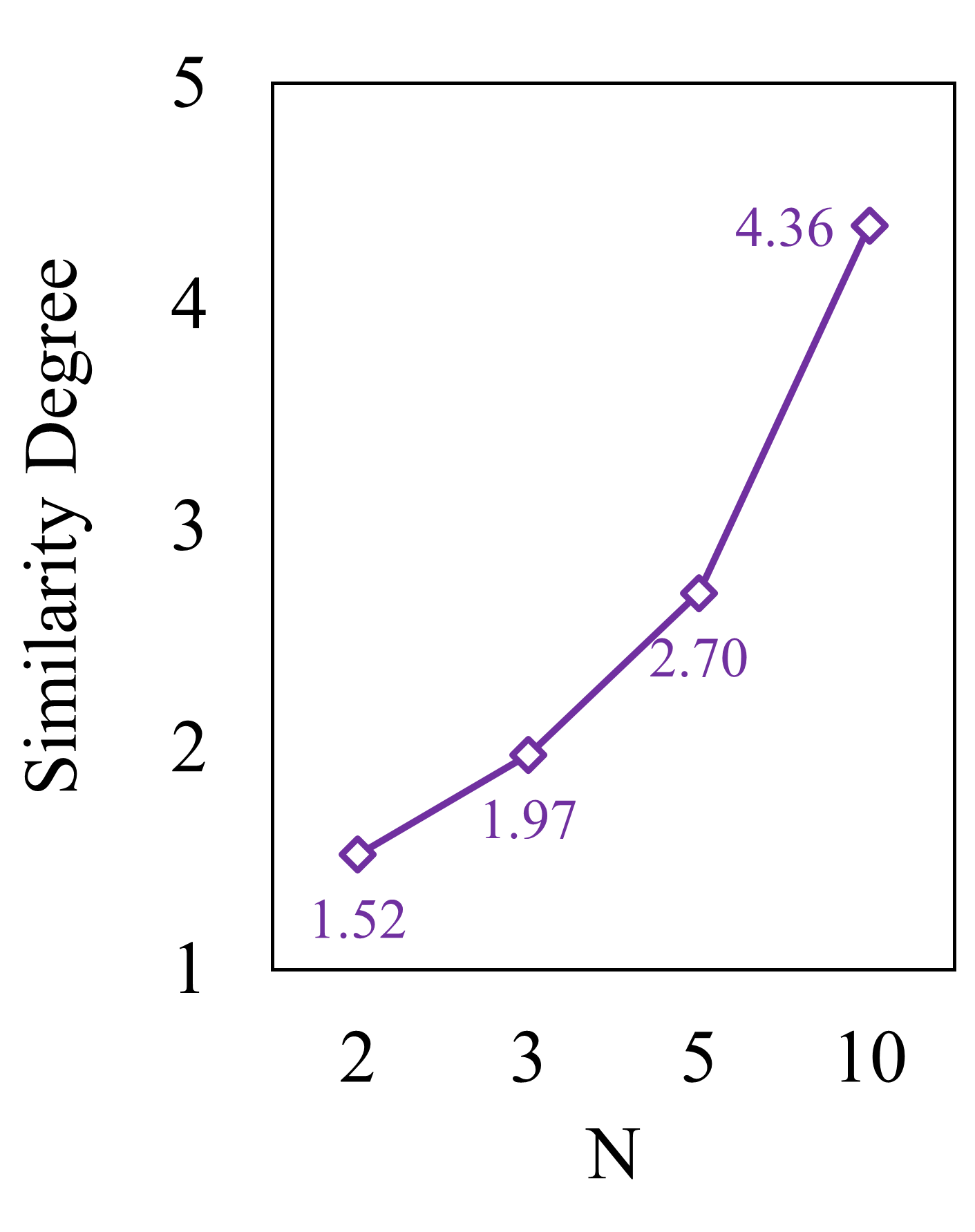}
    }
    \subfigure[Standard Deviation]{
    \includegraphics[width=0.245\textwidth]{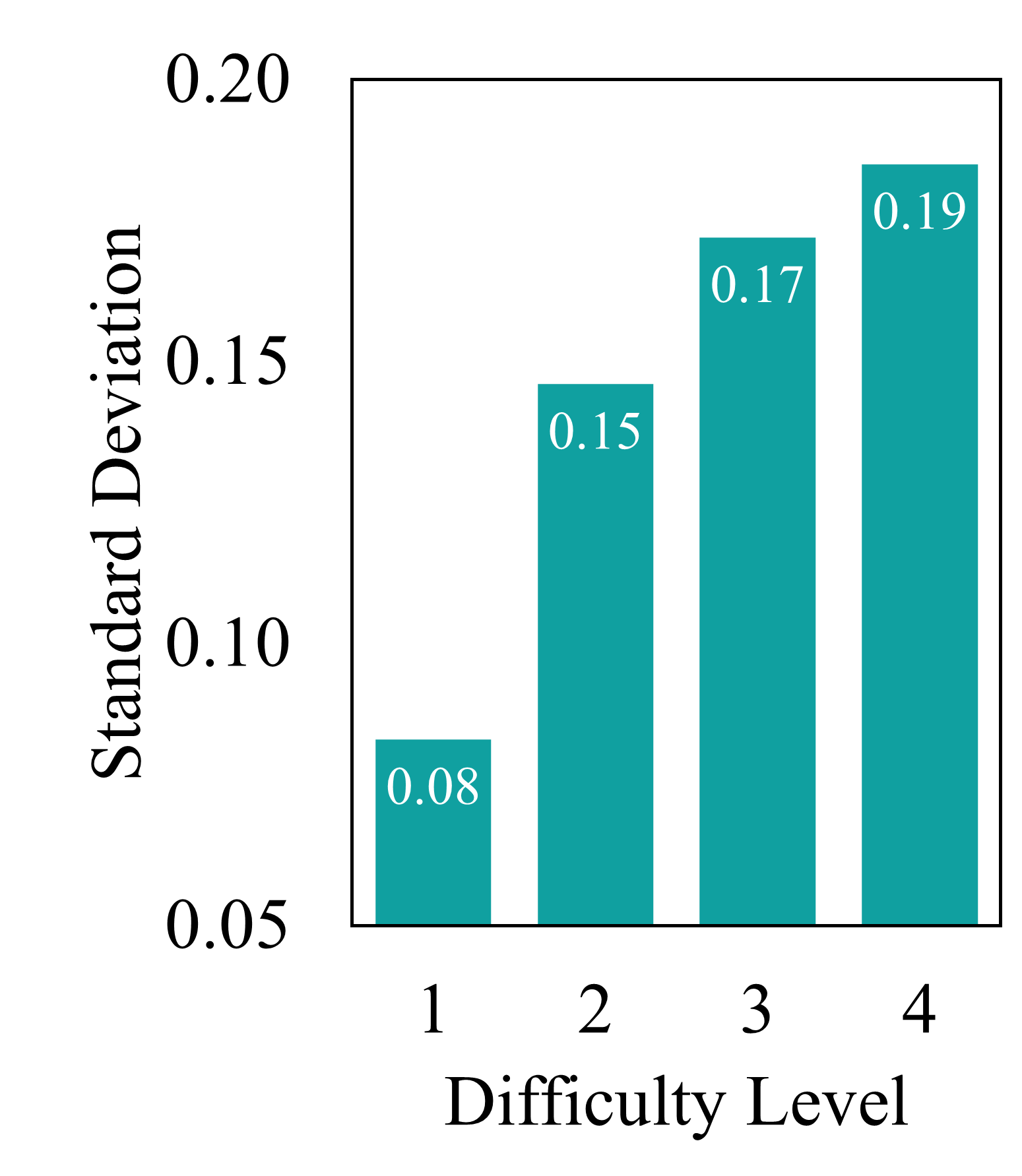}
    }
    \caption{Pilot experiment results using BFS on GSM8K, including: 
    (a \& b) Performances and costs when using different expansion size $N$, where the averaged number of generated token \#Token ($k$) is used to estimate the computational cost.
    (c) The averaged similarity degree of $N$ sub-nodes from 1000 randomly selected non-leaf nodes, which affects the severity of over-exploration.
    (d) Standard deviation of verifier scores for 10 sampled correct trajectories from each question at different difficulty levels. High-variance scores can lead to under-exploration of promising nodes.}
    \label{fig:pilot}
\end{figure}

\subsection{Results and Analyses}
\label{sec:pilot}

We adopt LLaMA-3-8B \citep{llama-3}, fine-tuned on the GSM8K training set, as the policy.
Following \citet{tian2024toward}, we construct a value network by equipping LLaMA-3-8B with a regression head to serve as the verifier for BFS.
Results on the GSM8K test set with different expansion sizes $N$ are shown in Fig.~\ref{fig:pilot}(a) and Fig.~\ref{fig:pilot}(b).

As $N$ increases, BFS consistently achieves better performance.
However, the computational costs rise dramatically. When comparing $N$\,$=$\,$5$ and $N$\,$=$\,$10$, there is a 3$\times$ to 9$\times$ increase in token consumption compared to greedy decoding, yet the improvement in accuracy is merely 0.5\%.
This raises concerns about the inefficiency of tree search algorithms in addressing search tasks within larger search spaces (i.e., larger $N$) for LLM reasoning.

A closer examination of the different problem difficulty categories in Fig.~\ref{fig:intro} reveals that the computational costs are quite similar across the categories, contradicting the intuition that resource allocation should increase with growing difficulty.
This indicates that current algorithms struggle to balance resource allocation, suffering from over-exploration in basic problems and under-exploration in complex cases.
Upon analyzing the constructed search trees, we discover two critical issues contributing to these problems.

\paragraph{Redundant Search States Result in Over-Exploration}
Due to the uncontrollable sampling of steps from the policy model, a large number of nodes convey identical semantic meanings (e.g., ``\textit{3+4=7}'' and ``\textit{4+3=7}'').
Expanding such nodes leads to redundant exploration.

To further investigate this, we randomly select 1,000 non-leaf nodes from BFS trees, expand $N=2,3,5,10$ sub-nodes for each, and apply agglomerative clustering on their vanilla SimCSE \citep{gao2021simcse} embeddings.
We then define the overall similarity degree for a group of sub-nodes as $\frac{N}{C}$, where $C$ represents the cluster number.
As shown in Fig.~\ref{fig:pilot}(c), the overall similarity degree increases linearly with the growth of $N$.
This indicates that the over-exploration issue can be more severe when taking a larger $N$ for higher accuracy.

\paragraph{High-Variance Verifier Scores Lead to Under-Exploration}
We observe that verifier scores often lack stability.
For instance, even though both trajectories lead to the correct answer, their scores may differ greatly.
As depicted in Fig.~\ref{fig:pilot}(d), we illustrate the standard deviation of scores for 10 sampled correct trajectories for each math problem across various difficulty levels, which are measured by the frequency of wrong answers within 64 sampled trajectories following \citet{wang2024litesearch}.
This highlights the high variance of verifier scores, especially for challenging problems.
Guided by inaccurate scores, the search process might be trapped in a locally optimal branch, leaving many valuable paths unexplored and resulting in their under-exploration.

This issue can be attributed to the high-variance training objective of the verifier produced by Monte Carlo (MC) rollouts, which has been thoroughly discussed in reinforcement learning theory~\citep{sutton2018reinforcement}.
Besides, due to the high costs of rolling out, the number of rollouts is often limited, which further increases the randomness of MC, especially for more difficult math problems where the policy is uncertain about the answer.


\section{\method{}}

We propose redundant state merging and score variance reduction to address the over-exploration and under-exploration issues analyzed in the pilot study.
Both methods are complementary and orthogonal with widely used tree search algorithms.

\begin{figure*}[t]
    \centering
    \includegraphics[width=0.99\textwidth]{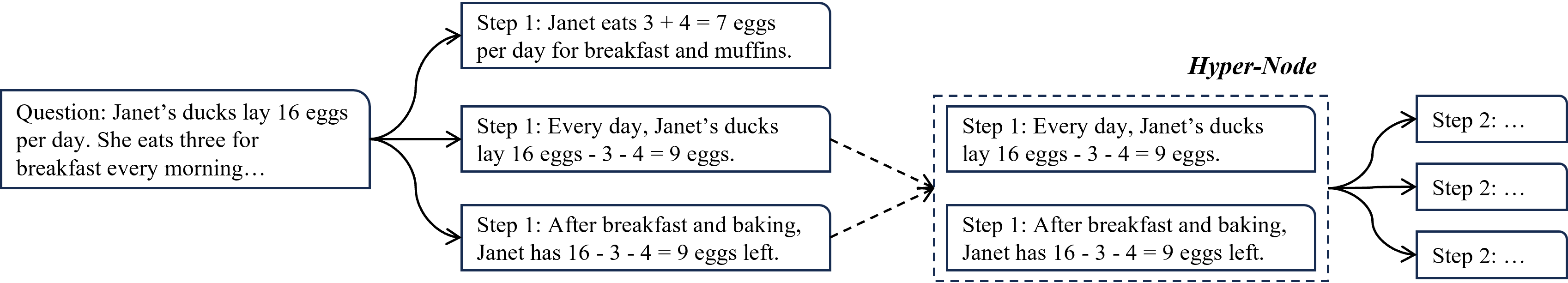}
    \caption{Illustration of redundant state merging. When new nodes are expanded, we merge semantically equivalent nodes into hyper-nodes using agglomerative clustering based on their embeddings.}
    \label{fig:method}
\end{figure*}

\subsection{Handling Over-Exploration via Redundant State Merging}
This method aims to merge semantically equivalent states to avoid over-exploration of them.
In the following subsection, we first introduce how to achieve this by utilizing a clustering algorithm and the step representations produced by an embedding model.
Then, we further investigate post-training the embedding model for better clustering.

\subsubsection{Semantic-Based State Merging}
As shown in Fig. \ref{fig:method}, state merging is implemented after each node expansion.
For newly created nodes $s_1,\dots,s_N$, we first employ an embedding model, such as SimCSE \citep{gao2021simcse}, to encode their reasoning processes.
Then, we utilize agglomerative clustering to partition them into $M$ groups:
\begin{equation}
\begin{aligned}
    \bar{\mathbf{s}}_1,\dots,\bar{\mathbf{s}}_M &= \mathtt{Agglomerative}(e_1,\dots,e_N),\\
    e_1,\dots,e_N &= \mathtt{Emb}(s_1),\dots,\mathtt{Emb}(s_N).
\end{aligned}
\end{equation}

In particular, we choose agglomerative clustering as it dynamically determines the cluster number based on the linkage distance threshold $d$ between clusters.
This offers a more flexible and adaptive solution, eliminating the need to pre-specify the number of clusters for accurate clustering.

Next, each group $\bar{\mathbf{s}}_i$ is used to create a new hyper-node, comprising multiple original nodes $\bar{\mathbf{s}}_i=\{ s^i_1,\dots,s^i_{|\bar{\mathbf{s}}_i|} \}$.
The corresponding verifier score $\bar{v}_i$ is calculated from the original nodes represented by the hyper-node:  $\bar{v}_i = f(v^i_1,\dots,v^i_{|\bar{\mathbf{s}}_i|})$, where $f$ can be $\mathtt{max(*)}$, $\mathtt{avg(*)}$, or $\mathtt{min(*)}$.

When expanding a hyper-node, we sequentially choose original nodes in the order of their verifier scores to obtain child nodes.
This facilitates the sampling of more diverse actions, thereby expanding the search space more effectively.

\subsubsection{Embedding Model Fine-Tuning}
Although we can directly adopt a pretrained embedding model for producing step embeddings, this results in sub-optimal clustering because of the domain discrepancy between the general corpus and target content.
For instance, we notice that SimCSE is often insensitive to numbers, which are critical for the math domain.
To address this issue, we propose two alternative data collection approaches to post-train the embedding model:
\textit{prompting-based} and \textit{consistency-based}.

\paragraph{Prompting-Based Approach}

This approach leverages the instruction-following capability of LLMs for semantic equivalence judgment.
Given reasoning states $s_i$ and $s_j$, we formulate the verification task as $\mathtt{LLM}(I, s_i, s_j)$, where $I$ denotes the prompt provided in Appendix \ref{sec:param}.
Though conceptually straightforward, its effectiveness hinges on the availability of a general LLM and a carefully engineered prompt.

\paragraph{Consistency-based Approach}
Inspired by the empirical finding that the same states often yield similar actions, we propose to craft labels for $s_i, s_j$ by comparing their impact on subsequent actions when swapped.
Formally, given $K$ rollouts $\mathbf{a}_1,\dots,\mathbf{a}_K$ sampled either from $s_i$ or $s_j$, we have
\begin{equation}
    \delta = \mathbb{E}_{k \in 1 \dots K}(|p(\mathbf{a}_k \mid s_i; \pi) - p(\mathbf{a}_k \mid s_j; \pi)|),
\end{equation}
where $\delta$ quantifies the influence of swapping $s_i$ and $s_j$ on rollout predictions of the policy $\pi$.
Intuitively, the smaller the $\delta$, the more similar the states are.
Thus, we consider $s_i$ and $s_j$ as identical when $\delta$\,$<$\,$\alpha$ and distinct when $\delta$\,$>$\,$\beta$, where $\alpha$ and $\beta$ are hyperparameters used to control the label quality.

After obtaining the labels of state pairs from the training corpus, we post-train the model using the standard cross-entropy loss:
\begin{equation}
    \mathcal{L} = y \log g(e_i, e_j) + (1 - y) \log (1 - g(e_i, e_j)),
\end{equation}
where $g(*,*)$ is the cosine similarity function and $y \in \{0,1\}$ denotes state equivalence.

\subsection{Handling Under-Exploration via Score Variance Reduction}
Previous studies \citep{wang2023math,tian2024toward} train their verifiers with inspiration drawn from the value function, which seeks to approximate the expected cumulative reward starting from a state $s$ and following a policy model $\pi$ thereafter.
This can be represented as \(v(s) = \mathbb{E}_{\pi} \left[ G_i \mid s_i = s \right]\), where \(G_i\) is the discounted return starting from state \(s_i\) and can be estimated using MC methods: $G_i \approx \frac{1}{\rho} \sum^{\rho}_{j=1} [ \mathbf{a}_j \text{ is correct}]$, where $\mathbf{a}_j \sim \pi(s_i)$ and $\rho$ is the number of rollouts.
As discussed in our pilot study, the high variance of verifiers trained in this way can mislead the search process, thus raising the under-exploration issue.
To address it, we propose leveraging \textit{TD($\lambda$)} and \textit{Ensembling} to reduce the score variance during training and inference, respectively.

\paragraph{Training Time}
TD($\lambda$) is a reinforcement learning algorithm blending key aspects from both MC methods and standard TD learning. 
In contrast to MC methods that rely on the outcome reward only, TD($\lambda$) balances immediate and delayed rewards by facilitating the allocation of credit for a reward back to previous states and actions.
Here, $\lambda$ refers to the trace decay parameter, with $0$\,$\leqslant$\,$\lambda$\,$\leqslant$\,$1$, allowing for a flexible trade-off between bias and variance in the estimates.
Formally, this process can be formulated as
\begin{equation}
    G_i^{\lambda} = (1 - \lambda) \sum_{t=1}^{T-i-1} {\lambda}^{t-1} G_i^{(t)} + \lambda^{T-i-1} G_i,
\end{equation}
where $T$ is the total number of steps and $G_i^{(t)}$ is equal to $v(s_{i+t})$.
Then, for state $s_i$, we train the verifier utilizing the standard mean-square error (MSE) loss:
\begin{equation}
    \mathcal{L}_{\mathrm{TD(\lambda)}} = (\min(G_i^{\lambda}, 1) - v(s_i))^2,
\end{equation}
where the $\lambda$-return $G_i^{\lambda}$ is constrained within the interval $[0,1]$ to satisfy the requirements of certain search algorithms.

\paragraph{Inference Time}
To further reduce the variance and bias arising from TD learning, we propose ensembling multiple verifiers during the inference time.
This approach can also be employed to avoid the need for training when open-source verifiers are accessible. 
Given $N_{vn}$ verifiers, the ensemble score for state $s$ is the average of their predictions: $\frac{1}{N_{vn}} \sum_{i=1}^{N_{vn}} v_i(s)$.
To avoid the impact on speed caused by running multiple verifiers serially, we deploy them in parallel on different devices.

\section{Experiment}

\begin{table*}[]
\small
\centering
\begin{tabular}{lccccccccc}
\toprule
& \multicolumn{2}{c}{GSM8K} & \multicolumn{2}{c}{GSM-Plus}  & \multicolumn{2}{c}{MATH} \\ 
\cmidrule(lr){2-3} \cmidrule(lr){4-5} \cmidrule(lr){6-7}
& Accuracy $\uparrow$ & \#Token ($k$) $\downarrow$ & Accuracy $\uparrow$ & \#Token ($k$) $\downarrow$ & Accuracy $\uparrow$ & \#Token ($k$) $\downarrow$ \\
\midrule
Greedy Decoding & .657 & 0.08 & .500 & 0.09 & .302 & 0.25 \\
Self-Consistency & .753 & 0.83 & .601 & 0.95 & .340 & 2.44 \\
Best-of-N & .819 & 0.83 & .634 & 0.95 & .352 & 2.44 \\
Weighted Voting & .817 & 0.83 & .620 & 0.95 & .372 & 2.44 \\
\cmidrule(lr){1-10}
BFS & .824 & 2.47 & .655 & 2.84 & .382 & 6.51 \\
~~~~w/ State Merge & .836 & \underline{0.88} & .664 & \underline{1.12} & .380 & \textbf{2.89} \\
~~~~w/ Var Reduce & \underline{.847} & 2.25 & \underline{.676} & 2.89 & \textbf{.388} & 7.63 \\
~~~~w/ \method{} & \textbf{.852} & \textbf{0.87} & \textbf{.684} & \textbf{1.09} & \underline{.384} & \underline{3.04} \\
\cmidrule(lr){1-10}
Beam Search & .828 & 1.55 & .669 & 1.88 & .378 & \underline{6.50} \\
~~~~w/ State Merge & .832 & \textbf{1.17} & .674 & \textbf{1.61} & .384 & \textbf{6.36} \\
~~~~w/ Var Reduce & \textbf{.846} & 1.58 & \underline{.680} & 1.89 & \underline{.386} & 7.45 \\
~~~~w/ \method{} & \underline{.842} & \underline{1.19} & \textbf{.682} & \underline{1.62} & \textbf{.396} & 7.04 \\
\cmidrule(lr){1-10}
Tree Search & .818 & 0.55 & .660 & 0.92 & .372 & 2.68 \\
~~~~w/ State Merge & .825 & \underline{0.48} & .657 & \textbf{0.69} & .378 & \textbf{1.58} \\
~~~~w/ Var Reduce & \textbf{.838} & 0.53 & \underline{.670} & 0.89 & \underline{.380} & 3.53 \\
~~~~w/ \method{} & \underline{.837} & \textbf{0.46} & \textbf{.677} & \textbf{0.69} & \textbf{.384} & \underline{2.00} \\
\cmidrule(lr){1-10}
MCTS & .838 & 10.7 & .676 & 14.1 & .374 & 59.0 \\
~~~~w/ State Merge & .845 & \underline{3.03} & .670 & \textbf{3.43} & .380 & \textbf{20.3} \\
~~~~w/ Var Reduce & \textbf{.854} & 8.84 & \underline{.686} & 14.4 & \underline{.394} & 66.6 \\
~~~~w/ \method{} & \underline{.853} & \textbf{2.60} & \textbf{.688} & \underline{3.53} & \textbf{.400} & \underline{22.9} \\
\bottomrule
\end{tabular}
\caption{Main test results. We emphasize the best results in \textbf{bold} and the second-best ones with \underline{underlining}.}
\label{tab:main}
\end{table*}

\subsection{Setup}

\paragraph{Datasets}
We conduct experiments on three popular test sets: GSM8K \citep{cobbe2021training}, GSM-Plus \citep{li2024gsm}, and MATH \citep{hendrycks2021measuring}.
GSM8K contains 1,319 grade school math problems taking between 2 and 8 steps to solve.
GSM-Plus is an enhancement of GSM8K by introducing 8 variations. We randomly sample one variant for each test case, resulting in 1,319 instances.
MATH consists of more challenging math problems from high school math competitions. Following previous work \citep{lightman2023let}, we test on a subset of 500 cases, named MATH500.

\paragraph{Models and Hyperparameters}
For GSM8K and GSM-Plus, the policy and value network are fine-tuned from LLaMA-3-8B \citep{llama-3} on the GSM8K training set.
For MATH, we employ the policy released by \citet{wang2023math}, which is based on Mistral-7B \citep{jiang2023mistral}, and adopt a PRM as the verifier due to its better performance on MATH.
Detailed parameter settings for training and search are reported in Appendix~\ref{sec:param} and \ref{sec:baseline}.

\paragraph{Baselines}
We report conventional greedy decoding and self-consistency \citep{wang2022self}, which only adopt the policy for inference.
Guided by verifiers, the sampling-based approach can be further enhanced using Best-of-N and Weighted Voting.
These methods select the solution with the highest score or ensemble the predicted answers by considering their respective scores as weights.
For tree search algorithms, we consider the most popular choices: Beam Search \citep{xie2024self,zhu2024deductive} and MCTS \citep{feng2023alphazero,tian2024toward}.
We also experiment with the typical BFS (\S \ref{sec:bfs}) and recent Tree Search \citep{wang2024litesearch}, which is an optimized implementation of A* \citep{nilsson1984shakey} that carefully manages the expansion budget to speed up the search process.

\paragraph{Evaluation Metrics}
We report the accuracy of answers (\textbf{Accuracy}) to evaluate performance and follow \citep{kang2024mindstar,wang2024litesearch} to adopt the averaged number of generated tokens (\textbf{\#Token ($k$)}) to estimate computational costs.

\subsection{Main Results}

Table \ref{tab:main} shows the main test results on the three datasets.
Generally, leveraging verifiers effectively improves performance over conventional methods relying only on the policy.
Besides, comparing these tree search methods, we can usually observe the performance gains when scaling the inference computation.
When applying our methods to these algorithms, we draw the following conclusions.

\paragraph{State Merging Effectively Decreases Computational Costs While Maintaining Performance}
State merging consistently reduces computational costs across all algorithms and datasets by enabling smarter node selection, which prevents redundant states from over-exploration.
For example, in BFS applied to GSM8K, state merging reduces \#Token~($k$) from 2.47 to 0.88, a nearly \textit{3$\times$ reduction} in computational overhead.
Beyond efficiency gains, we observe accuracy improvements in certain cases, even when fewer nodes are explored. This suggests that merging redundant states allows algorithms to allocate their computation budget more effectively, prioritizing promising paths over unproductive ones.

Among the four search algorithms tested, Beam Search shows the smallest token savings.
This stems from its inherent rigidity: it strictly expands a fixed number of nodes (determined by the beam size), limiting opportunities for optimization.
Nevertheless, Beam Search still benefits from state merging, as avoiding redundant node exploration enhances accuracy by redirecting resources toward more valuable states.  

\paragraph{Score Variance Reduction Substantially Improves Search Performance}
The application of score variance reduction leads to consistent performance gains across all tested algorithms, improving accuracy by approximately 1$\sim$2 points.
Notably, this method maintains stable token consumption on the GSM8K and GSM-Plus datasets, suggesting it efficiently allocates computational resources to valuable nodes rather than aimlessly switching trajectories, thereby overcoming the under-exploration issue.

However, on the more challenging MATH dataset, we observe a slight increase in computational costs.
This is likely due to the inherent difficulty of MATH problems, where search algorithms often struggle to converge on confident solutions.
To avoid under-exploration, our method would allocate more computations on unfinished trajectories as demonstrated by results in Fig.~\ref{fig:explore}.


\paragraph{\method{} Mitigates Over- and Under-Exploration}
By combining the two methods, we validate our approach enables more efficient and effective searching.
Furthermore, we observe better resource allocation for problems of varying difficulties, as demonstrated in Fig.~\ref{fig:explore}.
The results on both the GSM8K and MATH datasets indicate that our methods not only significantly reduce the token consumption for simple tasks but also ensure sufficient exploration of complex ones.
This strikes a balance between the problems of over-exploration and under-exploration, thereby enhancing the practicality of the tree search algorithms.

\subsection{Ablation Studies and Analyses}

In this part, we investigate the hyperparameter selection of our methods and conduct in-depth analyses of the working mechanism, mainly on the GSM8K dataset using the BFS algorithm.

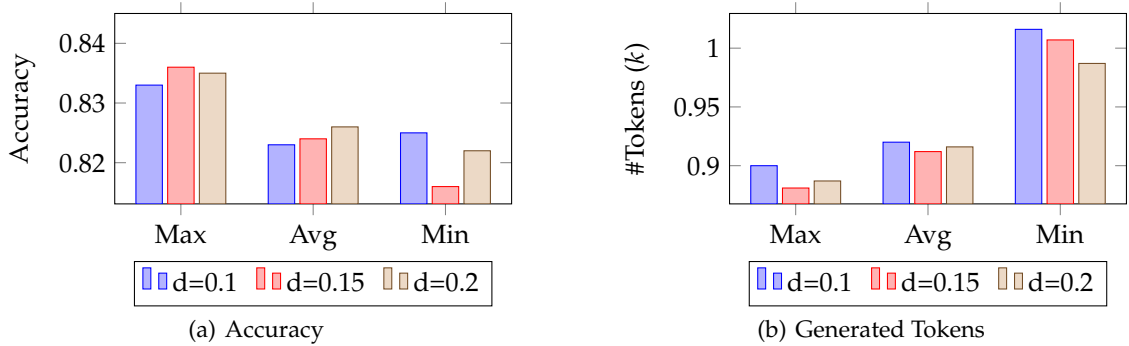
\begin{figure}[t]
\centering
\subfigure[Accuracy]{
\begin{tikzpicture}
    \begin{axis}[
        width=0.45\textwidth,
        height=0.27\textwidth,
        ybar,
        bar width=10pt,
        xmin=0.5,
        xmax=3.5,
        ymax=0.845,
        xtick={1,2,3},
        xticklabels={Max,Avg,Min},
        ylabel={Accuracy},
        legend style={at={(0.5,-0.3)}, anchor=north},
        legend columns=-1,
    ]
        \addplot coordinates {(1,0.833) (2,0.823) (3,0.825)};
        \addplot coordinates {(1,0.836) (2,0.824) (3,0.816)};
        \addplot coordinates {(1,0.835) (2,0.826) (3,0.822)};
        \legend{d=0.1\,\,\,, d=0.15\,\,\,, d=0.2}
    \end{axis}
\end{tikzpicture}
}
\hfill
\subfigure[Generated Tokens]{
\begin{tikzpicture}
    \begin{axis}[
        width=0.45\textwidth,
        height=0.27\textwidth,
        ybar,
        xmin=0.5,
        xmax=3.5,
        bar width=10pt,
        xtick={1,2,3},
        xticklabels={Max,Avg,Min},
        ylabel={\#Tokens ($k$)},
        legend style={at={(0.5,-0.3)}, anchor=north},
        legend columns=-1,
    ]
        \addplot coordinates {(1,0.900) (2,0.920) (3,1.016)};
        \addplot coordinates {(1,0.881) (2,0.912) (3,1.007)};
        \addplot coordinates {(1,0.887) (2,0.916) (3,0.987)};
        \legend{d=0.1\,\,\,, d=0.15\,\,\,, d=0.2}
    \end{axis}
\end{tikzpicture}
}
\caption{Ablation studies on parameter selection of $f$ and $d$ for state merging.}
\label{fig:merge_param}
\end{figure}

\begin{table}[]
\small
    \centering
    \begin{tabular}{l@{\hspace{3em}}cc}
    \toprule
    & Accuracy $\uparrow$ & \#Token ($k$) $\downarrow$ \\
    \midrule
    BFS (w/o state merge) & .824 & 2.47 \\
    \hdashline
    \multicolumn{3}{l}{BFS + Agglomerative} \\
    ~~~Pretrained SimCSE & .827 & 1.06 \\
    ~~~~~~w/ Prompt. FT & .835 & \textbf{0.88} \\
    ~~~~~~w/ Consist. FT & \textbf{.836} & \textbf{0.88} \\
    \hdashline
    \multicolumn{3}{l}{BFS + $k$-means (SimCSE w/ Consist. FT)} \\
    ~~~$k$=2 & .832 & 0.91 \\
    ~~~$k$=4 & .830 & 1.05 \\
    \bottomrule
    \end{tabular}
    \caption{Ablation studies of using different clustering algorithms and post-training strategies for state merging.}
    \label{tab:merge_ablation}
\end{table}

\paragraph{Influences of Hyper Parameter Selection for State Merging}
In Fig. $\ref{fig:merge_param}$, $d$ and $f$ determine the distances of nodes within each cluster during clustering and the verifier scores of the clustered hyper-nodes.
Increasing $d$ results in fewer clusters but higher probabilities of erroneously grouping different states into one cluster.
However, we notice our method is robust to this parameter.
For example when $f=\mathtt{max}(*)$, there is limited influence on both performance and efficiency by ranging $d$ from 0.1 to 0.2.
For $f$, the best choice is $\mathtt{max}(*)$ regarding both Accuracy and \#Token~($k$) among all examined values of $d$.
It might be because it better guarantees that the promising state (with a higher score) can be explored first.

In Table $\ref{tab:merge_ablation}$, we also compare different clustering algorithms.
We notice that agglomerative clustering works better than $k$-means.
Partial reason is that $k$-means requires specifying the number of clusters in advance, resulting in inflexible clustering and higher clustering error rates.
Nevertheless, $k$-means is also significantly better than the ``w/o state merge'' setting, proving the necessity of merging redundant states.

\paragraph{Influences of Post-Training Embedding Models}
We compare different strategies to post-train the embedding models in Table $\ref{tab:merge_ablation}$.
Prompting-based and consistency-based approaches achieve competitive performance, and both of them significantly outperforms the baseline, which only adopts the pretrained SimCSE for clustering.
To further validate these findings, we conduct a human evaluation to examine the correlation between similarity scores produced by these embedding models and human judgments in Appendix~\ref{sec:human_eval}.
Results in Table~\ref{tab:human_eval} again demonstrate the effectiveness of our post-training approaches.
Notably, the traditional rule-based method, edit distance, is very sensitive to dataset variations, limiting its robustness.
Besides, scaling the fundamental model size from RoBERTa-Base to Large yields measurable performance gains, suggesting future opportunities to advance state merging through more powerful embedding models.

\begin{table}[]
\small
    \centering
    \begin{tabular}{l@{\hspace{1.5em}}cc}
    \toprule
    & Accuracy $\uparrow$ & \#Token ($k$) $\downarrow$ \\
    \midrule
    \multicolumn{3}{l}{BFS + State Merge} \\
    ~~~~w/ MC & .836 & 0.88 \\
    ~~~~w/ TD($\lambda$) & .843 & 0.89 \\
    ~~~~w/ MC+Ensem & .848 & 0.88 \\
    ~~~~w/ TD($\lambda$)+Ensem & \textbf{.852} & \textbf{0.87} \\
    \bottomrule
    \end{tabular}
    \caption{Ablation studies of score variance reduction strategies.}
    \label{tab:value_ablation}
\end{table}

\paragraph{Comparison of Different Score Variance Reduction Approaches}
We observe that both TD($\lambda$) and verifier ensembling effectively enhance search performance (Table \ref{tab:value_ablation}) and improve verifiers for better step-level alignment with final accuracy (Fig. \ref{fig:value_calibration}).
Notably, ensembling shows better performance relative to TD($\lambda$) but at the cost of deploying multiple verifiers.
Combining these methods leads to further performance gains, demonstrating their complementary nature.
We also illustrate the score variance across problems of various difficulties in Fig.~\ref{fig:value_variance}, which again validates the effectiveness.

\paragraph{Inference-Time Scaling}

\begin{figure}
\centering
\begin{tikzpicture}[scale=0.8]
\begin{axis}[
    xlabel={\#Token ($k$)},
    ylabel={Accuracy},
    legend pos=south east,
    legend style={cells={anchor=west}},
]
\addplot
    coordinates {
        (0.250, 0.760)
        (0.449, 0.795)
        (0.982, 0.820)
        (2.473, 0.825)
    };
\addlegendentry{BFS} 

\addplot
    coordinates {
        (0.174, 0.748)
        (0.269, 0.780)
        (0.436, 0.816)
        (0.880, 0.836)
    };
\addlegendentry{BFS+State Merge} 

\addplot
    coordinates {
        (0.229, 0.773)
        (0.425, 0.818)
        (0.873, 0.843)
        (2.250, 0.847)
    };
\addlegendentry{BFS+Var Reduce} 

\addplot
    coordinates {
        (0.174, 0.741)
        (0.268, 0.800)
        (0.430, 0.825)
        (0.870, 0.852)
    };
\addlegendentry{BFS+\method{}} 

\end{axis}
\end{tikzpicture}
\caption{Inference-time scaling for the BFS algorithm equipped with our methods. The expansion budget $N$ is set as $2,3,5,10$.}
\label{fig:inference_scale}
\end{figure}
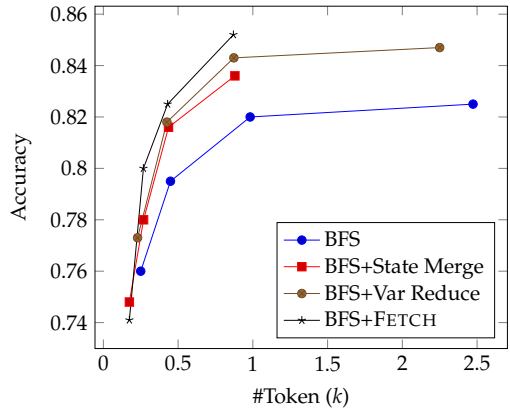

We validate the performance of our methods when scaling inference computations.
As shown in Fig.~\ref{fig:inference_scale}, our methods have a more rapid increase in Accuracy when increasing the expansion budget.
Besides, both state merging and variance reduction can lead to 1$\sim$2 times less computational costs with competitive performance.
Further combining both methods yields the best overall performance, again validating the conclusion of previous analyses.

\section{Related Work}

\paragraph{Deductive Reasoning with Search Algorithms}
LLMs have demonstrated remarkable potential in tackling complex reasoning tasks~\citep{achiam2023gpt,touvron2023llama,jiang2023mistral}.
However, they frequently fail on even simple reasoning steps, significantly degrading performance.
To mitigate this, researchers have proposed inference-time scaling methods that enhance computational effort during the reasoning process before finalizing answers.
Early approaches, such as Self-Consistency~\citep{wang2022self}, scaled computations by sampling multiple solutions.
This approach is inefficient because it requires exploring full solution paths, even if a mistake has occurred early on.
Later work introduced advanced tree search algorithms, including Beam Search~\citep{yao2024tree,zhu2024deductive,yu2024ovm}, MCTS~\citep{tian2024toward,zhang2024accessing}, and A*~\citep{wang2024q,wang2024litesearch}, to strategically allocate computational resources to critical reasoning steps, thereby improving efficiency.

Despite these advancements, most existing methods fail to address the challenges of over-exploration and under-exploration, leading to suboptimal efficiency.
To our knowledge, only \citet{tian2024toward} explicitly tackle over-exploration by pruning repeated reasoning states. However, their solution relies on either edit distance, which is prone to robustness issues (Table \ref{tab:human_eval}), or LLM-based similarity checks, which incur significant computational overhead.
This underscores the absence of a lightweight and robust mechanism to address this problem.

\paragraph{Training Verfiers to Guide Searching}
Previous studies have mainly utilized two types of verifiers: value networks \citep{yu2024ovm,tian2024toward} and process reward models (PRMs, \citealt{lightman2023let,wang2023math}).
Typically, both approaches leverage LLMs augmented with classification or regression heads to produce scalar scores to guide searching.

To collect training labels, \citet{lightman2023let} employed human annotators to identify reasoning errors in intermediate steps.
However, this approach is cost-prohibitive, hindering the scalability of training datasets.
Therefore, subsequent research \citep{yu2024ovm,wang2023math,tian2024toward,wang2024litesearch} has instead adopted MC methods to automate label collection.
While effective, they overlook the high variance associated with MC estimation, which influences the stability of trained verifiers and can lead to under-exploration.
In this study, we are the first to incorporate TD($\lambda$) and verifier ensembling to address this concern.


\section{Conclusion}

This work introduces \method{}, which addresses the over-exploration and under-exploration issues faced by tree search reasoning algorithms in LLMs by introducing two novel enhancements: state merging and variance reduction. By mitigating redundant search states and stabilizing verifier score estimation, these methods improve the practical feasibility of implementing advanced reasoning strategies, significantly enhancing accuracy and efficiency. Our experiments demonstrate the potential of these techniques to refine the deductive reasoning capabilities of LLMs without incurring prohibitive computational costs. Future explorations may further optimize these strategies and extend them to more complex tasks and domains.


\bibliography{ref}
\bibliographystyle{colm2024_conference}

\appendix

\section{Parameter Settings}
\label{sec:param}

For GSM8K, the policy model is fine-tuned from Llama-3-8B using standard cross-entropy loss.
We train the model for 2 epochs, adopting the AdamW optimizer with a linear scheduler using an initial learning rate of 5e-6.
The value network is trained on sampled rollouts from the policy, following previous studies \citep{yu2024ovm,tian2024toward}.
Specifically, for each question, we sample 16 rollouts with a temperature of 1 only from the initial state (i.e., $s_0=q$).
This model is trained for 1 epoch using typical mean square error loss with a learning rate of 2e-6.

For MATH, we directly adopt the policy model and PRM training data released by \citep{wang2023math}.
The policy model is based on Mistral-7B \citep{jiang2023mistral} trained from MetaMath \citep{yumetamath}.
We adopt the same value training script to train our PRM, and the trained PRM shows competitive performance as the official results \citep{wang2023math}.

When state merging, by default, the clustering parameters $d$ is 0.15 and $f$ uses $\mathtt{max}(*)$ across all experiments.
The embedding models are based on RoBERTa-Large \citep{liu2019roberta} pretrained using SimCSE \citep{gao2021simcse}.
For each dataset, we collect around 100k step pairs to post-train the embedding models using prompting-based or consistency-based methods.
The models are trained using binary cross-entropy loss with a learning rate of 1e-6.
For the prompting-based method, we adopt Llama-3-8B-Instruct \citep{llama-3}, which is trained on extensive post-training data, including publicly available instruction datasets and over 10 million human-annotated examples.
The instruction is designed as

\vspace{0.3em}
\noindent\fbox{%
    \parbox{0.97\linewidth}{%
        \emph{Given two reasoning steps (Step A and Step B) from two rationales of a math word problem, respectively. Your task is to judge whether these two steps are semantically equivalent.}\\
        \emph{Step A: $\{\tt{STEP_A}\}$}\\
        \emph{Step B: $\{\tt{STEP_B}\}$}\\
        \emph{Are Steps A and B semantically equivalent? Answer with Yes or No.}
    }%
}
\vspace{0.1em}

For the consistency-based method, we set $\alpha=0.02$, $\beta=0.08$, and $K=1,2$ for GSM8K and MATH, respectively.
By default, we adopt the consistency-based method as it does not rely on the other model.

As for score variance reduction, we adopt a combination of TD($\lambda$) and verifier ensembling for GSM8K, where $\lambda=0.8$ and $N_{nv}=2$.
We only use verifier ensembling with $N_{nv}=2$ for MATH because more accurate process rewards have been provided by \citep{wang2023math}, which are collected using very heavy MC sampling with enough rollouts for each intermediate state to reduce variance.

\section{Implementation of Search Algorithms}

\label{sec:baseline}

We report the parameter settings for the search algorithms used in this paper.
Except for greedy decoding, we set the generation temperature $\tau$ to 0.8 across all experiments.
We set the expansion budget $N=10$ for Self-Consistency, Best-of-N.
For BFS, we set $N=10$ and $N=5$ for GSM8K (or GSM-Plus) and MATH, respectively.
For BeamSearch, we set both the expansion budget and beam size to 5.
For Tree Search \citep{wang2024litesearch}, we set the maximum expansion budget $N=10$ and $N=5$ for GSM8K (or GSM-Plus) and MATH, and the expected accuracy $\epsilon=0.95$.
For MCTS \citep{tian2024toward}, we set the expansion budget $N=8$ for the root node and $N=4$ for the others. During simulation, we use a rollout number of 2.

\section{Human Evaluation on Embedding Models}
\label{sec:human_eval}

\begin{table}[]
\small
    \centering
    \begin{tabular}{lcc}
    \toprule
    & GSM8K & MATH \\
    \midrule
    Edit Distance & 0.307 & 0.500 \\
    \hdashline
    SimCSE (RoBERTa-Base) & 0.388 & 0.379 \\
    \hdashline
    SimCSE (RoBERTa-Large) & 0.464 & 0.414 \\
    ~~~~w/ Prompt. FT & 0.872 & \textbf{0.663} \\
    ~~~~w/ Consist. FT & \textbf{0.878} & 0.658 \\
    \bottomrule
    \end{tabular}
    \caption{Pearson Correlation Coefficient of similarity scores produced by different methods and human annotations on 200 sampled pairwise samples from GSM8K or MATH.}
    \label{tab:human_eval}
\end{table}

We randomly sample 200 step pairs from both GSM8K and MATH and hire annotators with natural language processing backgrounds to manually annotate whether they are the same steps with 0 or 1.
Each annotation is double-checked to ensure its correctness.
Then, we ask the embedding models to provide similarity scores for them and calculate the Pearson Correlation Coefficient (PCC) with human annotations.
As shown in Table \ref{tab:human_eval}, we again validate the effectiveness of post-training.
Besides, by comparing RoBERTa-Base and Large, we also demonstrate that scaling the model parameter may further enhance the model performance.
We leave it for future work.

\section{Analysis on Variance of Verifier Scores}
\label{sec:variance}

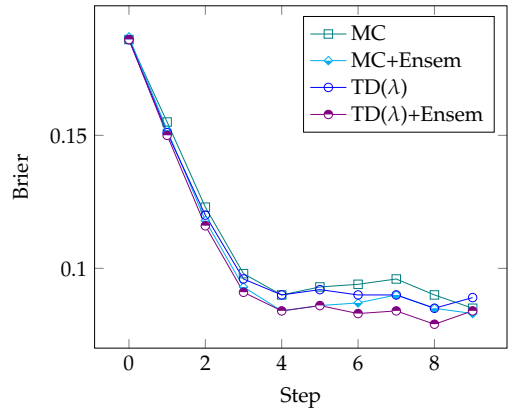
\begin{figure}
\centering
\begin{tikzpicture}[scale=0.8]
\begin{axis}[
    xlabel={Step},
    ylabel={Brier},
    ymin=0.07,
    ytick={0.1,0.15,0.2},
    legend pos=north east,
    legend style={cells={anchor=west}},
]
\addplot[
    color=teal,
    mark=square,
    ]
    coordinates {
    (0, 0.186)
    (1, 0.155)
    (2, 0.123)
    (3, 0.098)
    (4, 0.090)
    (5, 0.093)
    (6, 0.094)
    (7, 0.096)
    (8, 0.090)
    (9, 0.085)
    };
    \addlegendentry{MC}  
\addplot[
    color=cyan,
    mark=halfsquare*,
    ]
    coordinates {
    (0, 0.187)
    (1, 0.152)
    (2, 0.118)
    (3, 0.093)
    (4, 0.084)
    (5, 0.086)
    (6, 0.087)
    (7, 0.090)
    (8, 0.085)
    (9, 0.083)
    };
    \addlegendentry{MC+Ensem} 
\addplot[
    color=blue,
    mark=o,
    ]
    coordinates {
    (0, 0.186)
    (1, 0.151)
    (2, 0.120)
    (3, 0.096)
    (4, 0.090)
    (5, 0.092)
    (6, 0.090)
    (7, 0.090)
    (8, 0.085)
    (9, 0.089)
    };
    \addlegendentry{TD($\lambda$)} 
\addplot[
    color=violet,
    mark=halfcircle*,
    ]
    coordinates {
    (0, 0.186)
    (1, 0.150)
    (2, 0.116)
    (3, 0.091)
    (4, 0.084)
    (5, 0.086)
    (6, 0.083)
    (7, 0.084)
    (8, 0.079)
    (9, 0.084)
    };
    \addlegendentry{TD($\lambda$)+Ensem} 
\end{axis}
\end{tikzpicture}
\caption{Brier scores of estimated values from different verifiers at different reasoning steps and final accuracy, where brier scores can be calculated via mean square error (MSE).}
\label{fig:value_calibration}
\end{figure}

\begin{figure*}[t!]
    \centering
    \begin{tikzpicture}[scale=0.9]
    \begin{axis}[
        width=.8\textwidth,
        height=.2\textheight,
        xlabel={Difficulty Level},
        ylabel={Standard Deviation},
        ybar, bar width=15pt,
        xmin=0.5,xmax=4.5,
        xtick={1,2,3,4},
        legend style={cells={anchor=west}, at={(1.28, 0.67)}},
    ]
    \addplot
        coordinates {
            (1, 0.083)
            (2, 0.146)
            (3, 0.172)
            (4, 0.185)
        };
    \addplot
        coordinates {
            (1, 0.073)
            (2, 0.138)
            (3, 0.168)
            (4, 0.184)
        };
    \addplot
        coordinates {
            (1, 0.073)
            (2, 0.135)
            (3, 0.158)
            (4, 0.174)
        };
    \addplot
        coordinates {
            (1, 0.069)
            (2, 0.133)
            (3, 0.162)
            (4, 0.171)
        };
    \legend{MC, TD($\lambda$), MC+Ensem, TD($\lambda$)+Ensem}
    \end{axis}
    \end{tikzpicture}
    \caption{Comparison of the standard deviation of score estimation using different verifiers for 20 sampled reasoning paths of GSM8K questions at different difficulty levels.}
    \label{fig:value_variance}
\end{figure*}
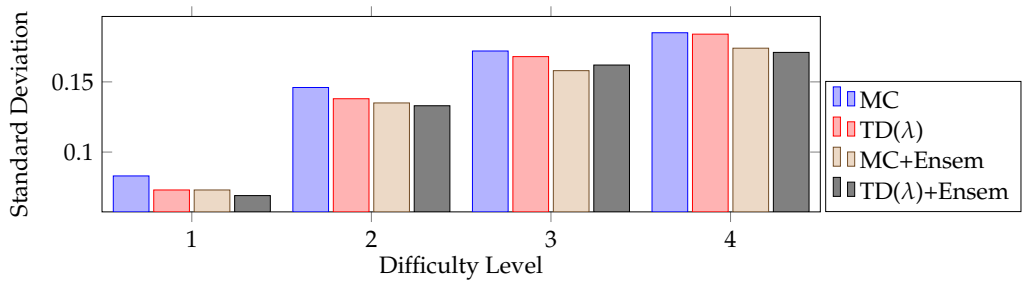

Fig. \ref{fig:value_variance} shows the variance of scores produced from verifiers trained in different methods.
We observe both TD($\lambda$) and verifier ensemble have lower variance than the baseline across subsets of various difficulties.
Their combination also makes further progress in most cases.
These results demonstrate that our methods can effectively reduce variance.

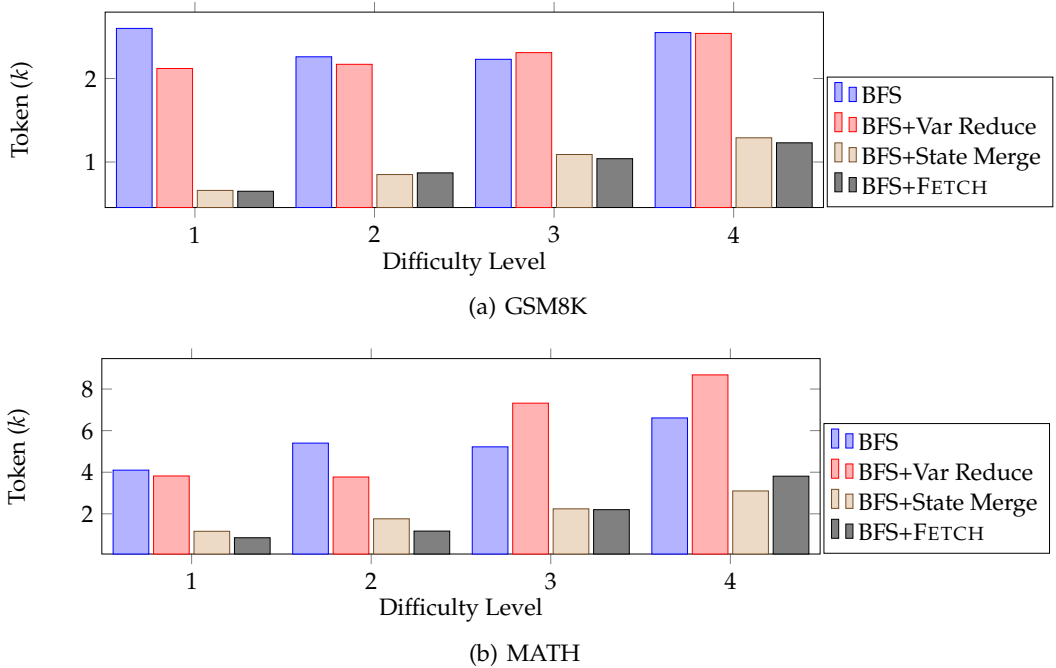
\begin{figure*}[t!]
    \centering
    \subfigure[GSM8K]{
    \begin{tikzpicture}[scale=0.9]
    \begin{axis}[
        width=.8\textwidth,
        height=.2\textheight,
        xlabel={Difficulty Level},
        ylabel={Token ($k$)},
        ybar, bar width=15pt,
        xmin=0.5,xmax=4.5,
        xtick={1,2,3,4},
        legend style={cells={anchor=west}, at={(1.32, 0.67)}},
    ]
    \addplot
        coordinates {
            (1, 2.60)
            (2, 2.26)
            (3, 2.23)
            (4, 2.55)
        };
    \addplot
        coordinates {
            (1, 2.12)
            (2, 2.17)
            (3, 2.31)
            (4, 2.54)
        };
    \addplot
        coordinates {
            (1, 0.66)
            (2, 0.85)
            (3, 1.09)
            (4, 1.29)
        };
    \addplot
        coordinates {
            (1, 0.65)
            (2, 0.87)
            (3, 1.04)
            (4, 1.23)
        };
    
    \legend{BFS, BFS+Var Reduce, BFS+State Merge, BFS+\method{}}
    \end{axis}
    \end{tikzpicture}
    }
    \subfigure[MATH]{
    \begin{tikzpicture}[scale=0.9]
    \begin{axis}[
        width=.8\textwidth,
        height=.2\textheight,
        xlabel={Difficulty Level},
        ylabel={Token ($k$)},
        ybar, bar width=15pt,
        xmin=0.5,xmax=4.5,
        xtick={1,2,3,4},
        legend style={cells={anchor=west}, at={(1.32, 0.67)}},
    ]
    \addplot
        coordinates {
            (1, 4.10)
            (2, 5.40)
            (3, 5.22)
            (4, 6.61)
        };
    \addplot
        coordinates {
            (1, 3.82)
            (2, 3.77)
            (3, 7.32)
            (4, 8.68)
        };
    \addplot
        coordinates {
            (1, 1.16)
            (2, 1.76)
            (3, 2.24)
            (4, 3.10)
        };
    \addplot
        coordinates {
            (1, 0.85)
            (2, 1.17)
            (3, 2.20)
            (4, 3.81)
        };
    
    \legend{BFS, BFS+Var Reduce, BFS+State Merge, BFS+\method{}}
    \end{axis}
    \end{tikzpicture}
    }
\caption{Ablation study of our methods on token consumption across GSM8K and MATH problems of different difficulties when using BFS.}
\label{fig:explore}
\end{figure*}

\end{document}